%% file: main.tex
\documentclass[letterpaper, 10 pt, conference]{ieeeconf}  

\IEEEoverridecommandlockouts                              
\usepackage[utf8]{inputenc}
\usepackage[T1]{fontenc}
\usepackage{microtype}
\usepackage[dvips]{graphicx}
\usepackage{xcolor}
\usepackage{times}
\usepackage{float}
\usepackage{amsmath}
\usepackage{amssymb}
\usepackage{mathabx}
\usepackage{dsfont}
\usepackage[ruled,vlined,noresetcount]{algorithm2e}
\usepackage{cancel}
\usepackage{multirow}
\usepackage{subcaption}
\usepackage{blindtext}
\usepackage{amssymb}
\usepackage{tikz}
\usetikzlibrary{shapes,arrows,calc,positioning}
\usepackage{makecell}
\usepackage{multirow}
\usepackage{array}
\usepackage{nicefrac}
\usepackage{graphicx}
\usepackage{todonotes}
\graphicspath{ {./images/} }

\usepackage{xcolor}
\usepackage{soul}

\usepackage[font=small]{caption}
\usepackage[colorlinks=true,linkcolor=black]{hyperref}

\title{\LARGE \bf
FollowMe: a Robust Person Following Framework Based on \\Visual Re-Identification and Gestures
}

\author{Federico~Rollo$^{\dagger,\ddagger,\S}$,
        Andrea~Zunino$^{\dagger,\ddagger}$,
        Gennaro~Raiola$^{\dagger,\ddagger}$,
        Fabio~Amadio$^{\dagger,\ddagger}$,\\
        Arash~Ajoudani$^{*,\ddagger}$,
        Nikolaos~Tsagarakis$^{*,\ddagger}$
\thanks{$^{\dagger}$Intelligent and Autonomous Systems, Leonardo Labs, Genoa, Italy}
\thanks{$^{\ddagger}$HHCM \& HRII, Istituto Italiano di Tecnologia, Genoa, Italy}
\thanks{$^{\S}$Industrial Innovation, DISI, Università di Trento, Trento, Italy}
\thanks{$^{*}$Equal advising}
\thanks{\footnotesize Authors' e-mails: {\tt\footnotesize\{name.surname\}.ext@leonardo.com}}
}

\begin{document}

\maketitle

\thispagestyle{empty}
\pagestyle{empty}

\begin{abstract}
    \input{sections/abstract}
\end{abstract}

\section{Introduction}
    \label{sec:intro}
    \input{sections/intro}

\section{Related Works}
    \label{sec:works}
    \input{sections/works}

\section{Person Perception}
    \label{sec:perception}

\input{sections/perception}

\input{sections/decisionmaking}

\section{Experiments}
    \label{sec:experiments}
    \input{sections/experiments}
    
\section{Conclusion}
    \label{sec:conclusion}
    \input{sections/conclusion}

\bibliographystyle{plain}
\bibliography{biblio}

\end{document}

%% file: sections/abstract.tex
Human-robot interaction (HRI) has become a crucial enabler in houses and industries for facilitating operational flexibility. When it comes to mobile collaborative robots, this flexibility can be further increased due to the autonomous mobility and navigation capacity of the robotic agents, expanding their workspace and consequently the personalizable assistance they can provide to the human operators. This however requires that the robot is capable of detecting and identifying the human counterpart in all stages of the collaborative task, and in particular while following a human in crowded workplaces. To respond to this need, we developed a unified perception and navigation framework, which enables the robot to identify and follow a target person using a combination of visual Re-Identification (Re-ID), hand gestures detection, and collision-free navigation. The Re-ID module can autonomously learn the features of a target person and uses the acquired knowledge to visually re-identify the target. The navigation stack is used to follow the target avoiding obstacles and other individuals in the environment. Experiments are conducted with few subjects in a  laboratory setting where some unknown dynamic obstacles are introduced. 

%% file: sections/intro.tex
The growing presence of robots in houses and industries is demanding an improvement in safety, robustness, and reliability for human-robot interactions and services. When a robot performs a collaborative task, it should have a robust environment understanding, while people should be free to move from one place to another without the need to manually command the robot's motions.
In this context, mobile robots are required to provide personal assistance; an important capability is to follow the human leader without being confused by other individuals in the robot’s perceived scene.
\begin{figure}
    \centering
    \includegraphics[width=1\linewidth]{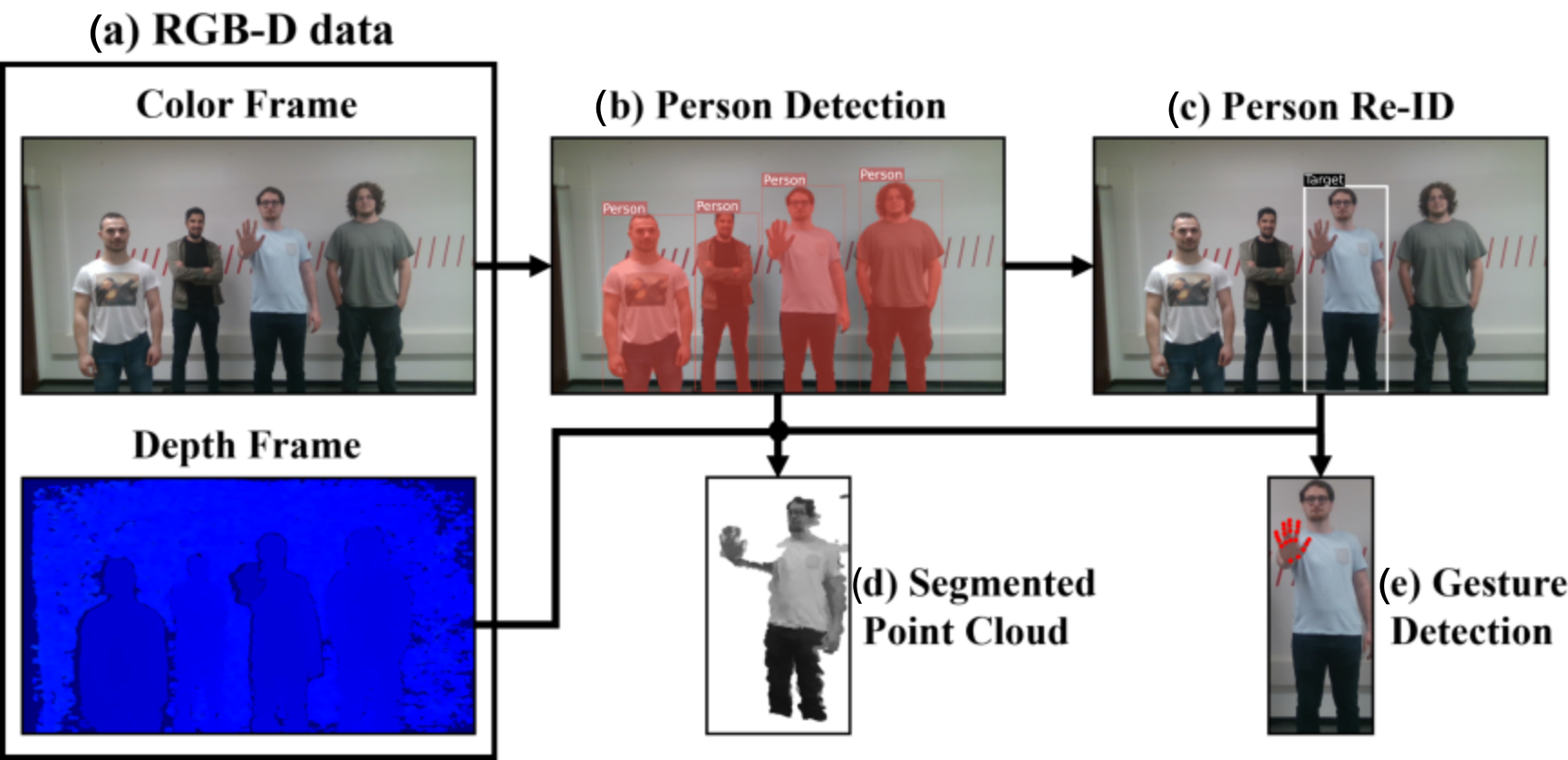}
    \caption{Perception pipeline: (a) RGB and Depth images acquisition; (b) person detection through Yolact++ (see Sect. \ref{subsec:person_detection}); (c) person Re-ID using deep neural network and human features distance (see Sect. \ref{subsec:reident}); (d) person localization using the point cloud and Re-Identified person mask; (e) gesture detection to send commands to the robot.}
    \label{fig:pipeline}
\end{figure}

In this work, we present a framework, \textit{i.e.} FollowMe, which performs a robust person-following task based on visual human Re-Identification (Re-ID) and hand gestures detection.
Our proposed architecture is composed of three modules: Perception, Decision Making, and Navigation, which are integrated with the robot through command references and sensor readings.
This framework\footnote{The code is available at \url{https://github.com/FedericoRollo/followme}} is adaptable to different floating base systems because it is based on established robotics tools and technologies such as ROS~\cite{quigley2009ros} and machine learning algorithms such as Yolact++~\cite{bolya2020yolact++}, MMT pre-training~\cite{ge2020mutual} and Mediapipe~\cite{zhang2020mediapipe}. 
The target person can move the robot to different locations without manual teleoperation. 
Visual Re-ID is a light, robust and personalised approach: the target does not have to wear clothes with specific patterns or other easily recognizable/intrusive devices, \textit{e.g.} IR emitters, to localize the target. We provide a module for hand gesture recognition to allow the target user to intuitively send commands to the robot (\textit{e.g.} stop). 

This work aims to enhance the assumption that human detection and Re-ID through visual data is a powerful tool for human-robot personalized collaboration. The application is user-friendly because the robot can autonomously collect and use calibration data to re-identify and follow the target person (see Sect.~\ref{subsec:reident}). In literature, such real complex problems are rarely treated; the research generally focuses on single topics (\textit{e.g.}, multi-object tracking) without considering mobile robotic integration in a real application environment which is not always trivial.
Among the possible applications for this framework, the most relevant are: carrying heavy items, moving the robot to different stations to perform manipulation, collaborative tasks \cite{lamon2020visuo}, and assisting people with care needs \cite{eisenbach2015user}.
The Re-ID and hand gesture detection modules are quantitatively evaluated with ad-hoc experimental setups showing impressive accuracy on test samples. The whole FollowMe framework is qualitatively evaluated in a simulated working area: the robot follows the target and actively responds to its hand gesture commands while avoiding the static and dynamic obstacles present in the area. 

The paper is structured as follows. State-of-the-art comparison is presented in Sect. \ref{sec:works}. The perception module comprising person detection and Re-ID, hand gestures detection, and 3D localization is explained in Sect. \ref{sec:perception}. In Sect. \ref{subsec:navigationstack} the navigation integration is presented while the decision-making module is introduced in Sect. \ref{subsec:statemachine}. The FollowMe experimental results and validation are reported in Sect. \ref{sec:experiments}. Finally Sect. \ref{sec:conclusion} draws conclusive remarks.

%% file: sections/works.tex
In industry and research, robots that perform person-following tasks already exist \textit{e.g.} the Piaggio Gita and Kilo\footnote{Piaggio Gita and Kilo: \href{https://www.piaggio.com/gb_EN/piaggio-world/about-us-piaggio/the-revolutionary-nature-of-gita-and-kilo/}{\nolinkurl{Gita-and-Kilo}}}. The existing applications use different kinds of sensors to accomplish this task. \cite{geetha2021follow} uses a Wi-Fi signal emitter and detector to identify the person to follow, while \cite{dang2011human} uses an IR camera to detect IR LEDs placed on the target. \cite{peng2016tracking} uses a sonar ring and a rangefinder and \cite{pradeep2017follow} uses a Bluetooth connection. Most of these frameworks need emitter/receiver devices to localize the target and usually, there is no integration with obstacle avoidance algorithms as opposed to \cite{afghani2013follow}, which uses an ultrasonic sensor to perform obstacle avoidance and an IR sensor to identify the LEDs attached to the person.
For robust detection, RGB-D cameras can be used. \cite{pinrath2018simulation} simulates a robot that uses a camera to detect a person with a red T-shirt and uses a range sensor to compute its distance from the robot and perform obstacle avoidance. \cite{shimoyama2017human} uses the shoulder length and the head height to distinguish persons and to re-identify the target one. \cite{sonoura2008person} fuses camera person detection with a laser range finder, while \cite{schlegel1998vision} uses colours and person contours for the following task. \cite{dalal2005histograms} uses holistic features (HOG) to detect persons. 
\cite{eisenbach2015user} has presented a person-caring robot for stroke patients in hospitals. They detect the legs with a laser range finder and the upper body with an orientation-based decision tree of HOGs. The outputs are merged with a person tracker and then the tracking is improved with colour and clothes texture matching. \cite{weber2017follow} performs the person-following task using an LSD-SLAM algorithm plus a CNN for head detection to reconstruct the 3D head surface. \cite{naseer2013followme} has developed a following drone; the quad-copter can fly and stabilize itself while maintaining the person in the camera field of view (FOV). The person is identified using holistic information (skeleton points) and some simple gestures (\textit{i.e.} based on the distance of the joints) are integrated to give commands. Recently, \cite{lamon2020visuo} has developed a visual-haptic application where a robot can track the person's skeleton using OpenPose \cite{cao2017realtime} and a stereo camera to project the 2D detection points in the 3D space. 

Most of the presented works lack robustness and cannot deliver personalised assistance, because they exploit information hardly distinguishable through individuals (e.g., location, pose, colours). 
Additional robust approaches are multi-object tracking (MOT) algorithms. TMOT~\cite{stadler2021improving} performs human tracking in cluttered and occluded environments but the time performance is not satisfactory for our goal (See Sect. \ref{sec:experiments}). Faster MOT algorithms~\cite{mostafa2022lmot} solve the timing issue, but tracking algorithms cannot reidentify objects that go out of view and then go back into the image plane, which is one of our strengths.

Our work generalizes the person-following task by detecting people on the image using deep neural network features to re-identify the target. The robot uses only visual data in the same way a human being would. Additional depth data are needed to accomplish the task. 
Our contribution resides in the development and integration of the whole visual pipeline: the visual person detection, re-identification, and localization modules plus the hand gestures detection for sending commands. This pipeline or a part of it can be simply re-adapted to other human-robot interaction tasks.

%% file: sections/perception.tex
We use visual and depth data obtained with an RGB-D camera, Fig.~\ref{fig:pipeline}a to perform four steps: $(i)$ Detection - Sect. \ref{subsec:person_detection}, $(ii)$ Re-ID - Sect. \ref{subsec:reident}, $(iii)$ Localization, and filtering - Sect. \ref{subsec:localization} and $(iv)$ Gesture detection - Sect. \ref{subsec:gesture}. The pipeline is displayed in Fig.~\ref{fig:pipeline}.

\subsection{Detection} \label{subsec:person_detection}
Instance Segmentation is a specific form of image segmentation that deals with detecting instances of objects and demarcating their boundaries. YOLACT++\cite{bolya2020yolact++} is a one-stage instance segmentation framework, which has some advantages over the existing architectures, one of the most significant ones is the speed of predictions. To the best of the authors' knowledge, YOLACT++ is the only existing framework that can deliver “real-time” instance segmentation inference\cite{bolya2020yolact++}. In our work, a standard model pre-trained on 80 classes of COCO~\cite{lin2014microsoft} (containing the class ``person") is adopted (Fig.~\ref{fig:pipeline}b). YOLACT++ can extract bounding boxes and segmentation masks belonging to the trained classes \textit{i.e.}, 1 for the image pixels where the object is localized, 0 around.
To precisely isolate the person's contours and fully remove the background information, the original images are element-wise multiplied by each binary mask. With this operation, the Re-ID module will be able to extract cleaner personal features based only on the individual due to the background noise removal. The resulting images, representing each detected person, are then available for the Re-ID module. 

\subsection{Re-Identification} \label{subsec:reident}
Person Re-ID is originally devised for the identification of a person across non-overlapping cameras.
Given a query image of a person, the goal is to recognize the same subject in images acquired by different cameras by analysing and extracting appearance information only (without using biometric cues).
For the Re-ID module, the popular Mutual Mean Teaching \cite{ge2020mutual} framework is used. In particular, only the pre-training phase is considered. In this phase, a deep learning network is trained to recognize people over images. The output images (see Sect. \ref{subsec:person_detection}) are passed inside the network to extract the features that will be used to re-identify the target person. Specifically, the features from the last layer of the trained network are considered. The Re-ID part is divided into two phases: a) Calibration and b) Identification. 
\begin{figure}
     \centering
     \includegraphics[width=0.8\linewidth]{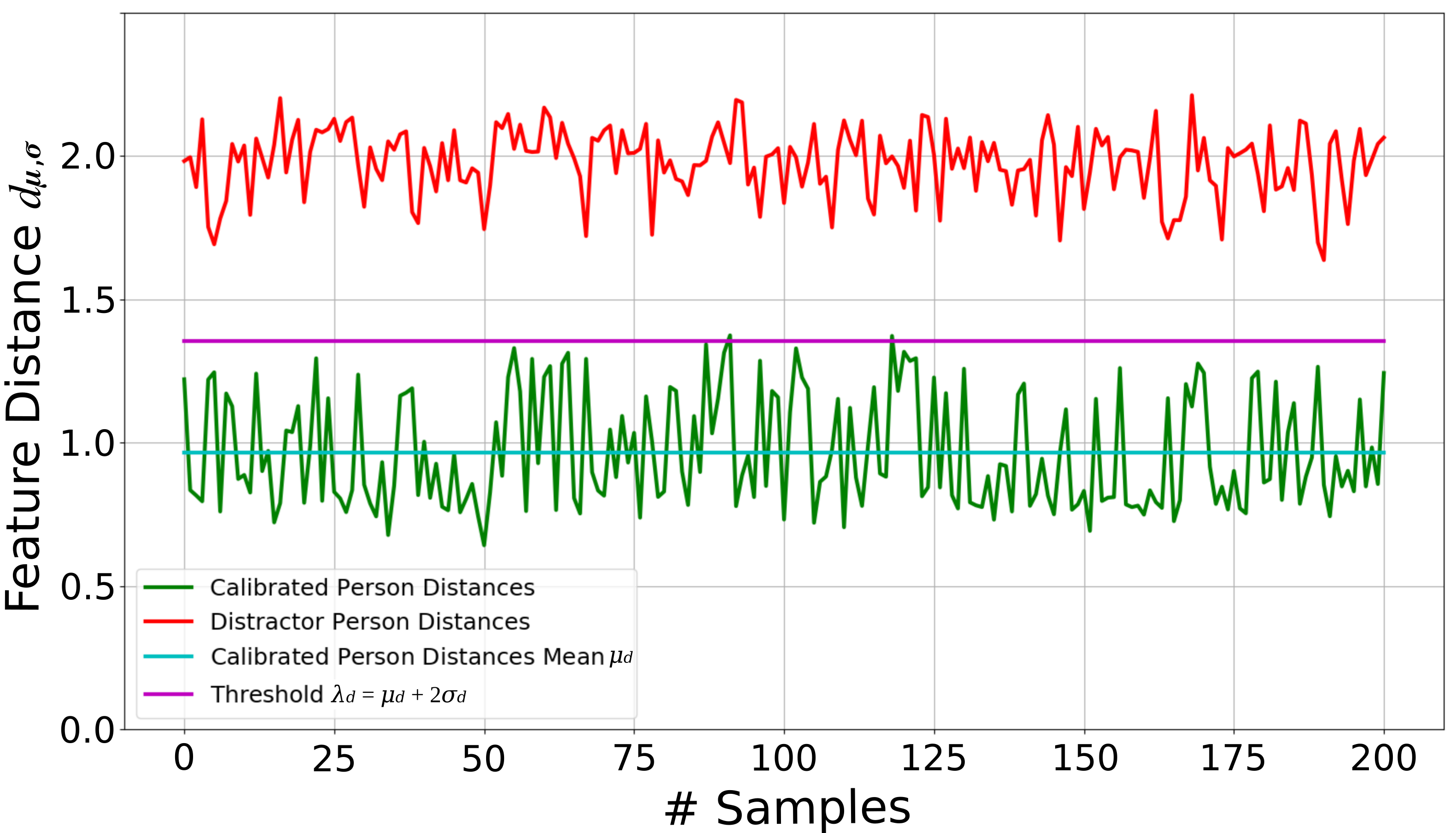}
     \caption{In green the plot of the feature distances for a calibrated person used to compute the target threshold (magenta horizontal line). In red the distances computed between the features of another person (the distractor) and the target template. The distances are computed over 200 sample images.}
     \label{fig:threshold}
\end{figure}
 
 \textbf{Calibration} At the beginning of the application, the target person is asked to move at different distances in front of the camera for a few seconds simulating the movements and postures which he/she will normally have while walking to collect images. The calibration step is crucial becuase the diversity in person representations during this stage affects the reidentification robustness to illumination changes and occlusions.
 Collected images are split into two groups, called \textit{person calibration} set (with $N_{c}$ elements) and \textit{threshold} set (with $N_\lambda$ elements).
 Following Sect. \ref{subsec:person_detection}, the processed images of the \textit{person calibration} set are given as input to YOLACT++ for the person masking and then to the Re-ID model for feature extraction.
 Let $D$ be the dimension of the feature vector $\mathbf{x}$ extracted by the Re-ID model, each person can be associated with a vector $\mathbf{x} = [x_1,\dots,x_D]^T$.
For each image of the \textit{person calibration} set the corresponding features are extracted and then used to compute the associated mean $\boldsymbol{\mu}=[\mu_1,\dots,\mu_D]^T$ and standard deviation $\boldsymbol{\sigma}=[\sigma_1,\dots,\sigma_D]^T$, which are used for target identification in the second phase.
 
Thus, given a new person image, the related feature $\mathbf{x}^*$ is computed and it is possible to identify him/her as the target person by measuring the following feature-space weighted distance between $\mathbf{x}^*$ and the distribution given by $\boldsymbol{\mu}$ and $\boldsymbol{\sigma}$:
 \begin{equation}\label{eq:feat_dist}
     d_{\boldsymbol{\mu},\boldsymbol{\sigma}}(\mathbf{x}^*) = \sqrt{\frac{1}{D}\sum_{i=1}^D \left(\frac{x^*_i-\mu_i}{\sigma_i}\right)^2}\text{,}
 \end{equation}
 
 where the subscript $i$ represents the $i$-th index of the corresponding vector. 
 The previously acquired \textit{threshold} set is exploited to compute a distance threshold $\lambda_d$: if $d_{\boldsymbol{\mu},\boldsymbol{\sigma}}(\mathbf{x}^*) \leq \lambda_d$ the person associated to feature $\mathbf{x}^*$ will be identified as the target seen during calibration.

Specifically, for each image in the \textit{threshold} set, the associated feature distance, Eq. \eqref{eq:feat_dist}, is computed, obtaining also the overall mean distance $\mu_d$ and standard deviation $\sigma_d$. The calibrated person threshold is set as  $\lambda_d=\mu_d+2\sigma_d$ to contain most of the samples without being too high to filter out distractor persons (every person different from the calibrated one is a distractor). Fig.~\ref{fig:threshold} reports an example set of distances for a calibrated person and a distractor, together with the relative $\lambda_d$ value.
In our work, a small number of wrongly re-identified targets (the false positives) is preferred because such cases are difficult to filter out while false negatives (calibrated person not re-identified) are simply discarded. This means that if the target is not recognized for some instants even if present on the image, the Kalman filter (see Sect. \ref{subsec:localization}) integrates the last detection for some time making the application more robust.

 \textbf{Identification} During the FollowMe application, new images are acquired, and the people are detected and filtered following Sect. \ref{subsec:person_detection}. As in the calibration phase, the background is removed from the processed images which are further passed inside the Re-ID model. The feature vectors obtained from the detected people Re-ID inference are collected. Finally, the feature distances (Eq. \ref{eq:feat_dist}) are computed. The person with the least distance, but less than the selected threshold $\lambda_d$, is selected as the target, Fig.~\ref{fig:pipeline}c. If every detected person has distances higher than the threshold, no target is detected.

\begin{figure}
    \centering
    \includegraphics[width=0.8\linewidth]{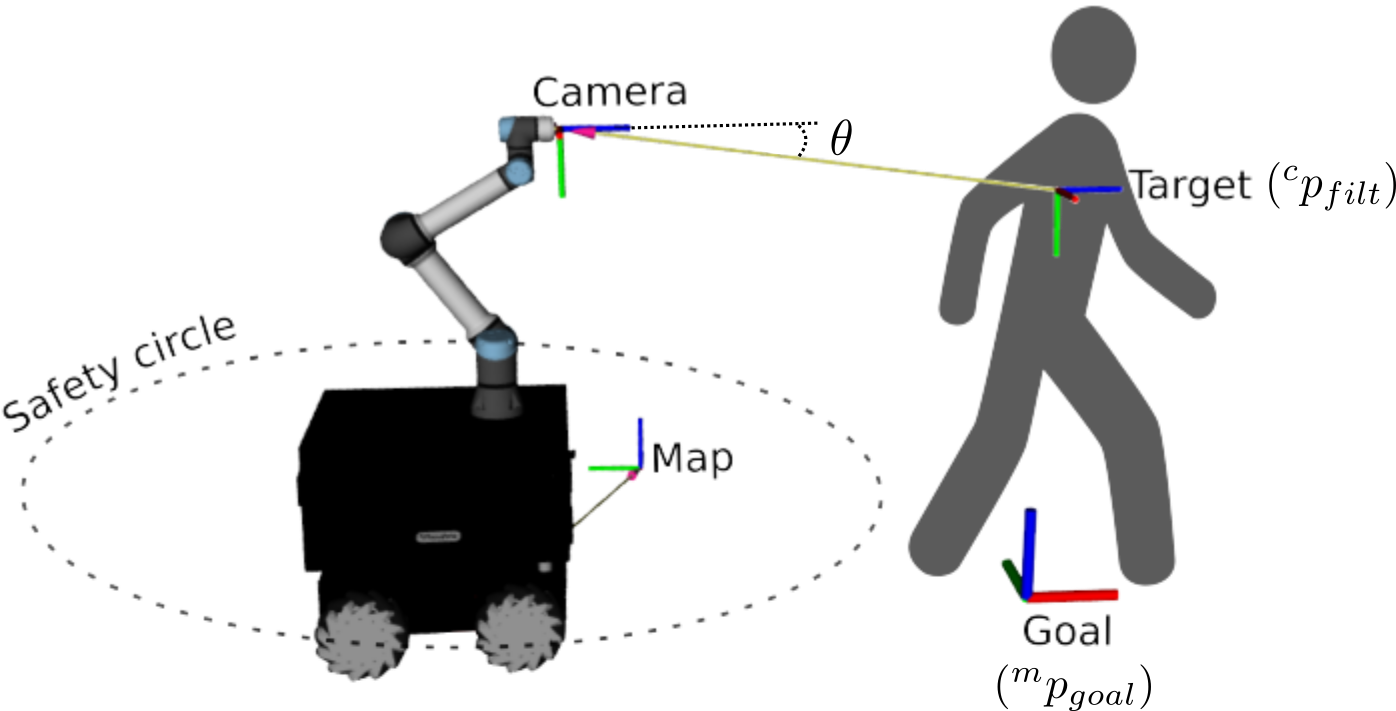}
    \caption{Overview of relevant transformation frames and representation of the safety circle used during the FollowMe application.}
    \label{fig:setup}
\end{figure}

\subsection{Localization and filtering} \label{subsec:localization}
The mask determined with the Re-ID step is used to compute the target position ${^c}p$ in the camera reference frame. Through RGB and depth data and thanks to the re-identified person mask, the target segmented point cloud is built, Fig.~\ref{fig:pipeline}d. The position ${^c}p$ is computed as the point cloud 3D centroid.\\ 
To filter out the noise caused by the robot's movements, vibrations and mask detection errors, a Kalman Filter (KF) is applied to ${^c}p$. The KF stops the state integration when measurement updates are not received for a predefined amount of time, the expiration time $t_{exp}$. The KF filtered position ${^c}p_{filt}$ is defined in the moving camera frame. For this reason, the homogenous transformation in Eq. \ref{eq:transformation} is applied:
\begin{equation} \label{eq:transformation}
    {^m}p_{filt} = {^m}T_c*{^c}p_{filt}\text{,}
\end{equation}

where ${^m}p_{filt}$ is the target position expressed in the map reference frame and ${^m}T_c$ is the transformation from the camera reference frame to the map one obtained from the navigation stack localization module (see Sect. \ref{subsec:navigationstack}).
Finally, the 2D goal position ${^m}p_{goal}$ is computed by projecting ${^m}p_{filt}$ on the 2D map plane (XY plane) and the heading angle $\theta$ between the camera reference frame and ${^c}p_{filt}$ is computed. The goal data (\textit{i.e.,} ${^m}p_{goal}$ and $\theta$) are managed by the state machine (see sec \ref{subsec:statemachine}) that will send them to the navigation module to drive the robot toward the goal. An overview of the relevant transformation frames is presented in Fig.~\ref{fig:setup}.

\subsection{Gesture detection} \label{subsec:gesture}

Hand gestures are a commonly used non-verbal communication tool to exchange information not only in robotics \cite{kaur2016review}. Using them it is possible to interact with the robot and convey commands to trigger the robot's functionalities and behaviours. In this work, gestures are used to stop and reactivate the FollowMe task but other commands can be added on necessity. 
The gesture detection module has been implemented using the Mediapipe hand tracking algorithm \cite{zhang2020mediapipe} and a Support Vector Machine (SVM) classifier on top. Mediapipe is a real-time framework, which extracts hand key points in 2.5 dimensions from a simple RGB image independently to hand scale, \textit{i.e.}, x,y and z position relative to the hand palm centre. Mediapipe inferences 21 ordered hand key-point with a high prediction quality and real-time inference, Fig.~\ref{fig:pipeline}e. 
To train the SVM classifier to distinguish gestures, The 21 hand landmarks were flattened in a single vector of size $63\ (21 \times 3)$.

\begin{figure}
    \centering
    \includegraphics[width=0.8\linewidth]{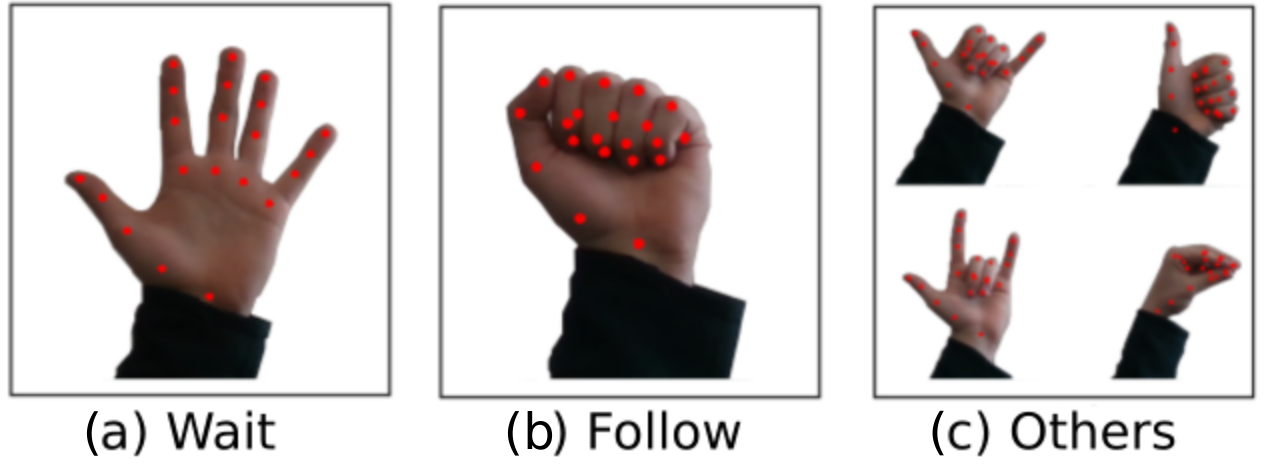}
    \caption{Classes considered to train the SVM for hand gesture detection. Mediapipe framework \cite{zhang2020mediapipe} is used to extract 3D key points (red points) relative to the centre of the hand.}
    \label{fig:handgestures}
\end{figure}

The classes representing the gestures are three: \textit{(i)} open hand for \textit{wait} command (Fig.~\ref{fig:handgestures}a), \textit{(ii)} closed hand for \textit{follow} command (Fig.~\ref{fig:handgestures}b), \textit{(iii)} all the other configurations to be ignored (Fig.~\ref{fig:handgestures}c).

A radial basis function kernel and a one-versus-one decision function were used to train the multi-class SVM classifier. 
The classification output is filtered (\textit{i.e.}, $\xi$ equal consecutive classification output are required to send a command, the default choice is $\xi=5$) and sent to the state machine, which handles it to command some robot actions (see Sect. \ref{subsec:statemachine}).

%% file: sections/decisionmaking.tex
\section{Navigation} \label{subsec:navigationstack}
The environments where humans and collaborative robots work are noisy and populated. The robot should navigate freely minimizing the collision probability with obstacles to prevent damage. To accomplish this task, we have used the ROS navigation stack\footnote{ROS Navigation Stack: \url{http://wiki.ros.org/navigation}}. Among the well-known advantages of using the navigation stack (\textit{e.g.}, customization, organization) there is also the generalization of this framework to a standard navigation architecture, which allows a simple re-adaptability to different robots. 

For safety reasons, a circular safety region is set around the robot (see Fig.~\ref{fig:setup}). If the position estimation of the target is inside this safety circle, \textit{i.e.} in a dangerous area, the robot deletes the last navigation goal and stops. Non-target persons are considered as obstacles and, if possible, they are avoided, otherwise, the robot stops and tries to find an alternative path through the goal.
In the worst case, \textit{i.e.} the target is not re-identified and the obstacle avoidance module doesn't recognize the person as an obstacle (a rare occurrence), the robot has the time to stop, due to Kalman filter expiration time, nullifying the chances of collision.
The safety distance $d_{safe}$ could be computed as follows:
\begin{equation} \label{eq:d_safe}
    d_{safe} = 1.4 * v_{max} * t_{exp}\text{,} 
\end{equation}
where $v_{max}$ is the max robot velocity and $t_{exp}$ is the Kalman filter expiration time. In this way, the robot should stop after a distance of $v_{max} * t_{exp}$ meters leaving an additional 40\% of the required distance from the person to ensure safety.

\section{Decision Making} \label{subsec:statemachine}
\begin{figure}[t]
    \centering
    \includegraphics[width=0.8\linewidth]{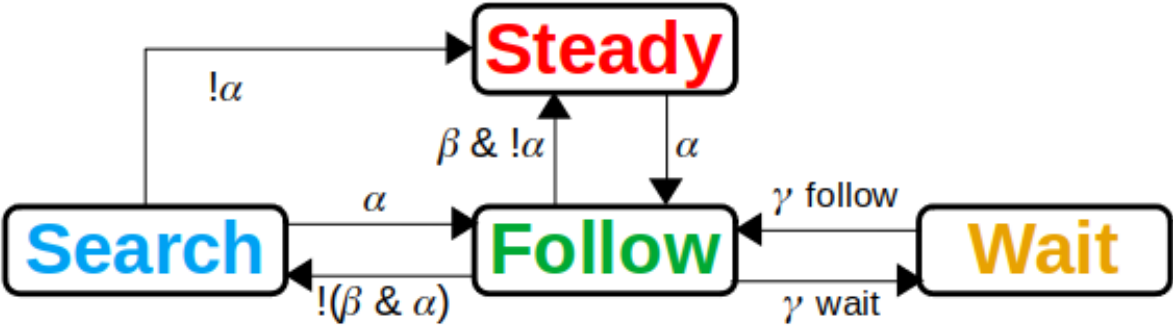}
    \caption{State machine diagram. The transitions between the states are highlighted with Greek letters and Boolean operators \textit{and} (\&), \textit{not} (!). The transitions' meaning is explained in Sect.~\ref{subsec:statemachine}.}
    \label{fig:statemachine}
\end{figure}

The application workflow is managed using a state machine, which collects all the data from the perception and navigation modules and controls the robot's behaviour.
The state machine has four states:
\begin{itemize}
    \item \textit{Steady}: the robot is stopped and waits from the perception module a \textit{follow} command.
    \item \textit{Follow}: the robot is following the target person.
    \item \textit{Search}: the robot, not receiving any target position, starts searching for him/her by rotating around itself (for at most one complete turn) toward the direction of the last detected position.
    \item \textit{Wait}: the robot has been stopped by the target person gesture \textit{wait} command and waits until the \textit{follow} gesture command is received.
\end{itemize}
In Fig.~\ref{fig:statemachine} the state machine is represented with a graph structure. The events which trigger the transitions (the edges) can be summarized as follows: $(\alpha)$ the target is re-identified in the camera FOV, $(\beta)$ the last target position is inside the safety circle and $(\gamma)$ the robot receives a command by the hand gestures module (Sect. \ref{subsec:gesture}).

%% file: sections/experiments.tex
We executed three different experiments to validate both individual modules (\textit{i.e.}, hand gesture classification - Sect.~\ref{subsec:gestureexperiment} and visual Re-ID - Sect.~\ref{subsec:reidentexperiment}) and the integrated system framework (Sect.~\ref{subsec:wholeexperiment}).

The experimental setup includes an Intel Realsense D415\footnote{Intel RealSense D415: \url{https://www.intelrealsense.com/depth-camera-d415/}} camera, a notebook with an \textit{Intel® Core™ i9-11950H} processor and an \textit{NVIDIA Geforce RTX 3080 Laptop} GPU and a Robotnik RB-Kairos+ 5e\footnote{RB-Kairos+ Mobile Manipulator: \url{https://robotnik.eu/products/mobile-manipulators/rb-kairos/\#more-versions}} as the assistant robot. The camera was fixed on the UR5e robot wrist with a 3D printed adapter and the robotic arm was positioned vertically to point the camera forward (see Fig.~\ref{fig:setup}). The safety circle is set to $1.25 m$ following Eq. \ref{eq:d_safe} with $v_{max} = 0.3 m/s$ and $t_{exp} = 3 s$. The navigation relies on an Adaptive Monte Carlo Localization algorithm\cite{xiaoyu2018adaptive} to estimate the robot's position using odometry and laser scan data.  
Regarding the Re-ID experiments, we pre-trained an IBN-ResNet-50~\cite{pan2018two} on the popular MSMT17~\cite{wei2018person} dataset and set the feature dimension of the network $D$ to 256 which is a popular choice in Re-ID settings because it is a good trade-off between performances and feature lengths.

\begin{center}
    \begin{table}[h]
        \centering
        \caption{Classification metrics over a test set of images. on the left, hand gestures, and on the right, Re-ID.}
        \begin{tabular}{|c|ccc||cc|}
            \hline
                    & \multicolumn{3}{c||}{Hand Gestures}  & \multicolumn{2}{c|}{Re-ID} \\
            \hline
                    & \textit{Wait} & \textit{Follow}    & \textit{Others} & \textit{Target} & \textit{No Target} \\
            \hline
            \textbf{Precision}    & 0.98      & 0.99      & 0.96  & 0.96       & 0.91   \\
            \textbf{Recall}   & 0.99      & 0.96      & 0.97   & 0.91      & 0.96 \\
            \textbf{F1-score}  & 0.98      & 0.97      & 0.96 &  0.93      & 0.93  \\
           \hline\hline
             \textbf{Accuracy} & \multicolumn{3}{c||}{0.97}  & \multicolumn{2}{c|}{0.94}\\
             \hline
        \end{tabular}
        \label{tab:metrics}
    \end{table}
\end{center}

\subsection{Hand gesture classification} \label{subsec:gestureexperiment}
To train the SVM classifier, a dataset of 400 images was acquired with the hand in different positions. 
A standard classification validation was used with a group of 8 persons for the experimental process to validate this module. For each person and each class (\textit{i.e.}, \textit{Wait}, \textit{Follow}, \textit{Others}), the hand key points were extracted from 500 images for a total of 4000 images for each class (i.e. a total of 12000 images). These data were classified with the SVM. Using the results and the ground truth labels, the mean (over the subjects) confusion matrix  and some classification metrics were computed and presented in Fig.~\ref{fig:handconfusionmatrix} and the left part of Table \ref{tab:metrics}, respectively. The presented numbers showed the power of the algorithm in distinguishing the gestures among a variety of different actors.

\subsection{Re-identification} \label{subsec:reidentexperiment}

\begin{figure} [t]
    \centering
     \begin{subfigure}{0.49\linewidth}
         \centering
         \includegraphics[width=\linewidth, trim={0cm 0cm 0cm 0cm}]{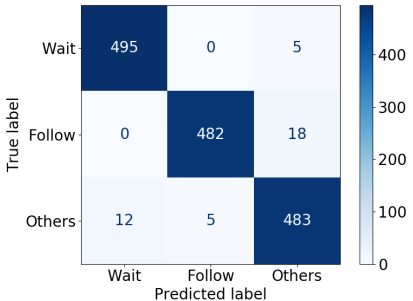}
         \caption{Average confusion matrix for hand gesture over a test set of images.}
         \label{fig:handconfusionmatrix}
     \end{subfigure}
     \begin{subfigure}{0.49\linewidth}
         \centering
         \includegraphics[width=\linewidth, trim={0cm 0cm 0cm 0cm}]{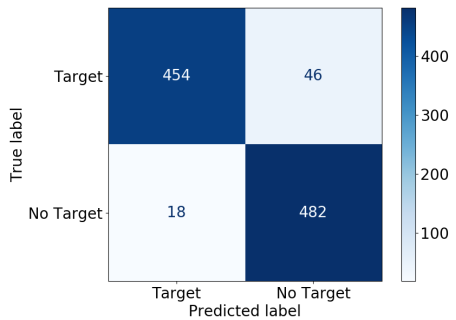}
         \caption{Confusion matrix for Re-ID over a test set of images.}
         \label{fig:reident_matrix}
     \end{subfigure}
    \caption{Comparison between panoptic and instance segmentation inferences.}
    \label{fig:panopticsegm}
\end{figure}

To validate the Re-ID module, a dataset of 8500 images was collected. 500 images with a single subject for each person were taken ($500 * 8 = 4000$ images) For calibration. The remaining 4500 images were collected according to the following logic: 500 images in which all the persons are present and 500 images for each person in which all the persons but the person of interest are included. The person Re-ID module is calibrated with the 500 calibration images where $\frac{2}{3}$ are used for the calibration and $\frac{1}{3}$ for the threshold computation. Finally, for each person, 1000 testing images were used and divided into two sets: the one where all the persons are present and the one where all persons but the calibrated one are present. The classification was evaluated as follows: for the first set, the classification is correct if the module correctly re-identifies the target, \textit{i.e.} it does not re-identifies another person or no one. for the second set, where the calibrated person is not present, the classification is considered correct if and only if the module does not re-identify anyone \textit{i.e.} all the feature distances are above the threshold. The result of such a process was represented in the confusion matrix in Fig.~\ref{fig:reident_matrix} and the right part of Table \ref{tab:metrics}. The numbers are averaged per subject \textit{i.e.} one subject is used for calibration and the others as distractors, alternating for each subject and finally averaging the results. The numeric results confirm that the chosen threshold computation favours the false negatives (top-right values in Fig.~\ref{fig:reident_matrix}) but it penalizes the false positives (bottom-left values in Fig.~\ref{fig:reident_matrix}) which is precisely the desired behaviour. This can be stated also from the \textit{Target} column of Table \ref{tab:metrics} where precision is higher than recall. 

\subsection{FollowMe} \label{subsec:wholeexperiment}

\begin{figure}[t]
    \centering
    \includegraphics[width=0.8\linewidth, trim={0cm 0cm 0cm 0cm}]{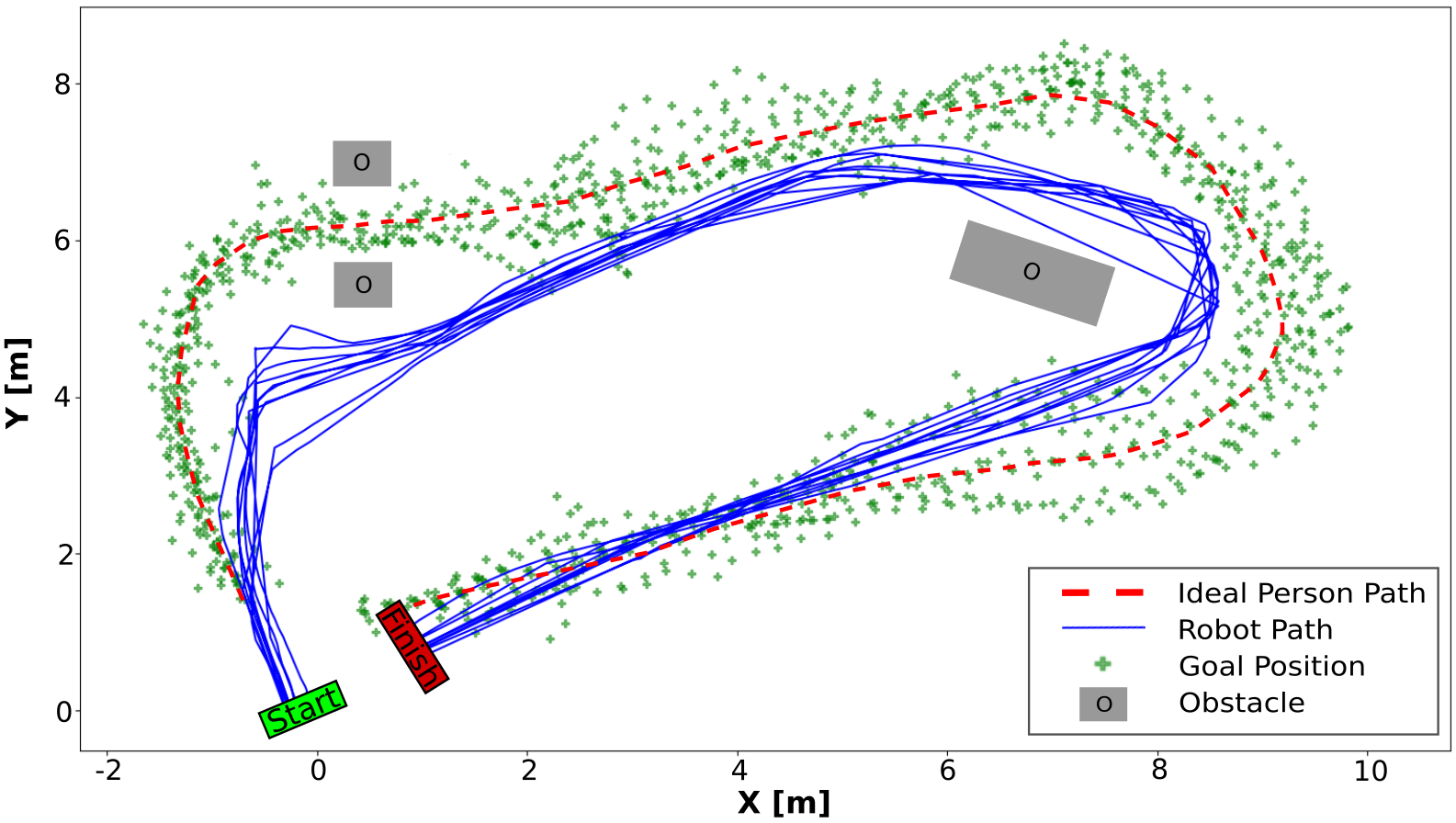}
    \caption{Results of the FollowMe experiment: each person has to follow an ideal path (red dashed line) while the robot (blue line) has to follow him/her. The goal positions computed from perception module data are represented with green plus signs. The robot is placed in the green start position and has to follow the target until the red finish position is reached.}
    \vspace{-0.5cm}
    \label{fig:path_chart}
\end{figure}

The whole framework was validated with a group of 10 persons. The robot starts with a fixed starting position in a laboratory area (approximately $100m^2$) where its empty map is known and some random unknown obstacles are placed. Each person has to follow a predefined path while other persons are moving around (generally, no more than 4 persons are present along the path). At the start, the robot waits for a \textit{Follow} gesture command and, when the target reaches the final station, the robot has to be stopped using the \textit{Wait} hand gestures. The experiment was performed one time for each person and the qualitative results are shown in Fig.~\ref{fig:path_chart}. The robot (blue path) detects the person's position (green $+$) and follows him/her avoiding unknown obstacles and maintaining a pattern path similar to the target ideal one (dashed red line).
Some videos presenting the FollowMe application are available online\footnote{FollowMe Youtube videos: \url{https://www.youtube.com/playlist?list=PLdibjJfM06zvKRkrR0fo7UNAxMfaa9h_E}}.

We computed the time computation of our overall proposed framework to have a time performance estimation. The time frequency, from camera acquisition to robot goal sending, ranges from $10Hz$ to $7Hz$ depending on the number of persons present in the image (from 1 to 10). Our system runs online, slowing down as the number of detected persons increases. However, in a real-applied scenario, it is rare to detect more than 10 not occluded persons.

%% file: sections/conclusion.tex
This work presented a robust framework for following a target person by a mobile robot, mainly based on visual Re-ID and gesture detection. The experiments confirmed that human Re-ID is a strong feature to perform person-following tasks and suggested that this ability could be crucial for other human-centred applications. Using a simple and not invasive RGB-D camera the robot can efficiently track a selected person simplifying the human-robot collaboration. The FollowMe framework is easily customizable to a target person, avoiding mismatches or switches with other persons, \textit{i.e.} the distractors, even when two persons are similarly dressed.
Despite this, during experiments, some limitations were faced. In specific light conditions, \textit{i.e.,} when a strong light is pointed towards the camera, the detection or Re-ID module lacks robustness. This limitation is also connected to the specific camera hardware. Also, Re-ID is strictly correlated to online calibration, \textit{i.e.,} if the calibration is not properly finalized, the further recognition of the target could be less robust; the robot should see the person at different distances and in different configurations during calibration, simulating as much as possible the human motion made at inference time.

Future directions refer to \textit{(i)} introduce lidar information to track the person out of the camera FOV, \textit{(ii)} predict the direction of the target and \textit{(iii)} replace the calibration phase with a continual learning approach.